\pdfoutput=1
\documentclass[11pt]{article}
\usepackage[final]{acl}
\usepackage{tcolorbox}
\usepackage{times}
\usepackage{latexsym}
\usepackage{booktabs}
\usepackage{graphicx}
\usepackage{amsmath}
\usepackage{multirow}
\usepackage{pifont}
\usepackage{xspace}
\usepackage{amssymb}
\usepackage{hyperref}
\usepackage{natbib}
\usepackage{graphicx}
\usepackage{enumitem}
\usepackage{pgfplots}
\usepackage{tabularx}
\usepackage{caption}
\usepackage{subcaption}
\usepackage{bbm}
\usepackage{url}
\usepackage{amsmath} 
\usepackage{pifont}
\usepackage{arydshln}
\usepackage{makecell}
\usepackage{booktabs}
\usepackage{multirow} 
\usepackage{algorithm}
\usepackage{algorithmic}

\usepackage{amsmath}
\usepackage{fontawesome5}

\usepackage[T1]{fontenc}

\usepackage[utf8]{inputenc}
\usepackage{microtype}

\usepackage{xcolor}

\DeclareUnicodeCharacter{211D}{\ensuremath{\mathbb{R}}}
\DeclareUnicodeCharacter{2124}{\ensuremath{\mathbb{Z}}}
\DeclareUnicodeCharacter{21A6}{\ensuremath{\mapsto}}

\DeclareUnicodeCharacter{2264}{\ensuremath{\leq}}
\DeclareUnicodeCharacter{2265}{\ensuremath{\geq}}
\DeclareUnicodeCharacter{2192}{\ensuremath{\rightarrow}}
\DeclareUnicodeCharacter{2194}{\ensuremath{\leftrightarrow}}
\DeclareUnicodeCharacter{2227}{\ensuremath{\wedge}}
\DeclareUnicodeCharacter{22A2}{\ensuremath{\vdash}}
\DeclareUnicodeCharacter{2080}{\ensuremath{_0}}
\DeclareUnicodeCharacter{2081}{\ensuremath{_1}}
\DeclareUnicodeCharacter{2082}{\ensuremath{_2}}
\DeclareUnicodeCharacter{2083}{\ensuremath{_3}}
\DeclareUnicodeCharacter{2084}{\ensuremath{_4}}
\DeclareUnicodeCharacter{2085}{\ensuremath{_5}}
\DeclareUnicodeCharacter{2086}{\ensuremath{_6}}
\DeclareUnicodeCharacter{2087}{\ensuremath{_7}}
\DeclareUnicodeCharacter{2088}{\ensuremath{_8}}
\DeclareUnicodeCharacter{2089}{\ensuremath{_9}}
\DeclareUnicodeCharacter{2200}{\ensuremath{\forall}}
\DeclareUnicodeCharacter{2203}{\ensuremath{\exists}}
\DeclareUnicodeCharacter{2228}{\ensuremath{\lor}}
\DeclareUnicodeCharacter{2260}{\ensuremath{\neq}}
\DeclareUnicodeCharacter{230B}{\ensuremath{\rfloor}} 
\DeclareUnicodeCharacter{2223}{\ensuremath{\mid}}    
\DeclareUnicodeCharacter{230A}{\ensuremath{\lfloor}} 
\DeclareUnicodeCharacter{2090}{\ensuremath{_a}}  
\DeclareUnicodeCharacter{2091}{\ensuremath{_e}}  
\DeclareUnicodeCharacter{2092}{\ensuremath{_o}}  
\DeclareUnicodeCharacter{2093}{\ensuremath{_x}}  
\DeclareUnicodeCharacter{2211}{\ensuremath{\sum}}        
\DeclareUnicodeCharacter{2115}{\ensuremath{\mathbb{N}}}  

\DeclareUnicodeCharacter{2208}{\ensuremath{\in}} 

\title{Verified Critical Step Optimization for LLM Agents}

\author{%
Mukai Li$^{\dag,1,2}$ \quad Qingcheng Zeng$^{1,3}$ \quad Tianqing Fang$^{\dag,1}$ \quad
Zhenwen Liang$^{1}$\quad \\
\textbf{Linfeng Song}$^{1}$  \quad
\vspace{-5pt}
\textbf{Qi Liu}$^{\dag,2}$\quad
\textbf{Haitao Mi}$^1$\quad
\textbf{Dong Yu}$^{1}$
\vspace{10pt}
\\
$^1$Tencent AI Lab \quad $^2$The University of Hong Kong \quad  $^3$Northwestern University\\
\texttt{kaikiaia3@gmail.com},\ \texttt{fangtq229@gmail.com},\ \texttt{liuqi@cs.hku.hk}  \\
\small{
\faGithub ~\url{https://github.com/kiaia/CSO}}  \\
\small{
\faGithub ~\url{https://github.com/Tencent/CognitiveKernel-Pro}} 
}

\begin{document}
\maketitle


\begin{abstract}
As large language model agents tackle increasingly complex long-horizon tasks, effective post-training becomes critical. Prior work faces fundamental challenges: outcome-only rewards fail to precisely attribute credit to intermediate steps, estimated step-level rewards introduce systematic noise, and Monte Carlo sampling approaches for step reward estimation incur prohibitive computational cost. Inspired by findings that only a small fraction of high-entropy tokens drive effective RL for reasoning, we propose \textbf{Critical Step Optimization (CSO)}, which focuses preference learning on \emph{verified critical steps}, decision points where alternate actions demonstrably flip task outcomes from failure to success. Crucially, our method starts from failed policy trajectories rather than expert demonstrations, directly targeting the policy model's weaknesses. We use a process reward model (PRM) to identify candidate critical steps, leverage expert models to propose high-quality alternatives, then continue execution from these alternatives using the policy model itself until task completion. Only alternatives that the policy successfully executes to correct outcomes are verified and used as DPO training data, ensuring both quality and policy reachability. This yields fine-grained, verifiable supervision at critical decisions while avoiding trajectory-level coarseness and step-level noise. Experiments on GAIA-Text-103 and XBench-DeepSearch show that CSO achieves 37\% and 26\% relative improvement over the SFT baseline and substantially outperforms other post-training methods, while requiring supervision at only 16\% of trajectory steps. This demonstrates the effectiveness of selective verification-based learning for agent post-training.
\end{abstract}
\section{Introduction}

\begin{figure}[t]
\centering
\includegraphics[width=0.9\columnwidth]{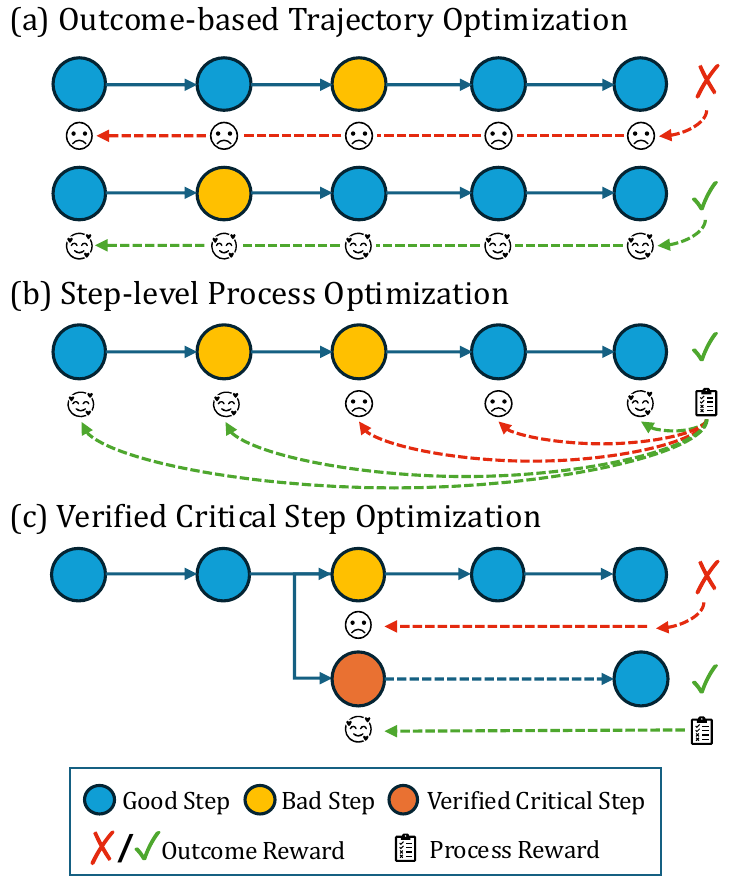}
\caption{\textbf{Comparing post-training paradigms for LLM agents.}
(a) Trajectory-level optimization assigns coarse outcome rewards to entire trajectories, while (b) step-level process optimization relies on intermediate rewards which could be inaccurate. (c) Verified Critical Step Optimization (CSO) focuses learning on verified critical steps where alternative actions demonstrably flip task outcomes, providing precise credit assignment with minimal supervision.}
\label{fig:teaser}
\end{figure}
The rapid advancement of large language models (LLMs), especially those tailored for multi-step reasoning and decision-making, such as GPT-5~\citep{openai_o1_2024},  Claude 4.5~\citep{claude4.5}, Kimi-K2~\citep{kimiteam2025kimik2openagentic}, DeepSeek-R1~\citep{2025deepseekr1} and DeepSeek-V3.2~\citep{deepseekv3.2}, has laid the foundation for increasingly capable agentic systems that solve complex tasks through iterative tool use and environment interaction. As these LLM-based agents tackle more long-horizon, multi-stage problems, effective post-training techniques become critical for translating general model capabilities into robust and aligned behavior in deployment scenarios. Recent pipelines of agent post-training~\citep{li2025websailor, wu2025webdancer, wan2026deepverifier, hu2025webcot} adopt a two-stage paradigm: supervised fine-tuning (SFT) followed by reinforcement learning (RL). 
However, SFT often suffers from off-policy distribution shift~\citep{tao2025webshaperagenticallydatasynthesizing,wang2025exploreevolvescalingevolved}. 
While on-policy RL mitigates this, it requires high computational cost from full rollouts and suffers from sparse and delayed reward signals that hinder credit assignment~\citep{wu2025webdancer,wei2025webagentr1trainingwebagents, wang2026spposequencelevelppolonghorizon}.

Recent work jumping out of the scope of SFT and online RL has explored \emph{offline or semi-online post-training} approaches that balance efficiency and performance. As shown in Figure~\ref{fig:teaser}, one line focuses on \emph{outcome-based trajectory optimization}. Methods like ETO~\citep{song2024eto} and MiroThinker~\citep{mirothinker} build preference pairs from successful and failed trajectories, offering scalability but suffering from imprecise credit assignment in which reasonable actions in failed trajectories are penalized uniformly, while suboptimal decisions in successful ones are reinforced.
Another line pursues \emph{step-level process optimization}. AgentRPM~\citep{choudhury2025agentprm} trains process reward models for intermediate assessments, while Co-Evolving Agents~\citep{jung2025coevolving} uses failure agents to generate hard negatives. Though finer-grained, these methods rely on estimated process rewards that introduce noise and systematic biases. 
Hybrid approaches like IPR~\citep{xiong2024ipr} attempt to combine both paradigms but face significant practical challenges. IPR derives step-level rewards through Monte Carlo sampling from each step forward, which becomes prohibitively expensive on complex tasks. 
From another perspective, existing methods overlook that \emph{not all steps are equally important}, as many involve trivial actions or few alternatives. Recent work on RLVR, particularly entropy-oriented algorithms~\citep{wang2025beyond}, shows that only a small fraction of high-entropy tokens drive effective RL for reasoning. This suggests focusing on \emph{critical branching points} where alternate actions drastically change outcomes, rather than uniform or dense supervision.

We propose \textbf{Critical Step Optimization (CSO)}, which focuses preference learning on \emph{verified critical steps} where alternate actions change the task outcome from failure to success. Unlike trajectory-level methods that apply coarse rewards uniformly, CSO provides fine-grained, step-specific supervision. Unlike step-level methods relying on estimated rewards, CSO grounds supervision in verified outcomes, eliminating process reward noise. Inspired by the high-entropy token principle in~\citet{wang2025beyond}, we identify critical steps as pivotal decisions (e.g., tool selection, query formulation) that determine trajectory success.
Starting from \emph{failed policy trajectories}, we use a PRM to efficiently identify candidate critical steps where the policy errs but expert alternatives show promise. We verify these candidates through branch rollouts: replacing the policy's action with an expert alternative and continuing with the policy itself until completion. Only verified successful branches become preference pairs, contrasting the successful alternative against the original failed action. This yields verifiable, fine-grained preference data that achieves semi-on-policy coverage, offline training efficiency, and outcome-verified supervision.

We evaluate on GAIA~\citep{mialon2023gaia} and XBench-DeepSearch~\citep{chen2025xbench} using CK-Pro-8B~\citep{fang2025cog}. Our contributions are threefold: \textbf{(1)} We identify limitations of existing approaches and propose CSO, which focuses on verified critical steps where alternative actions demonstrably flip task outcomes. \textbf{(2)} CSO achieves 37\% and 26\% relative gains over SFT baseline on GAIA-Text-103 and XBench-DeepSearch, enabling an 8B model to match GPT-4.1, while requiring supervision at only 16\% of trajectory steps and outperforming all baseline methods. \textbf{(3)} We demonstrate that combining PRM identification with outcome verification effectively pinpoints critical decision steps in agent execution, achieving superior performance by focusing learning on these pivotal branches that determine task success.
\section{Related Work}

\subsection{Preference-based Post-training of Agentic Models}
While online reinforcement learning from verifiable rewards (RLVR) and self-evolving paradigms have demonstrated significant utility across diverse domains \cite{2025deepseekr1, dong2025agentic_arpo, li2025websailor, yu2026rfew, huang2025rzero, fang2025webevolver}, preference-pair-based learning remains a highly effective and widely adopted strategy for agent post-training~\citep{mirothinker}. 
Building on base LLMs, recent research has explored various strategies to bolster agentic capabilities. For instance, \citet{webthinker} integrate a Deep Web Explorer into a think-search-draft loop, utilizing DPO with human feedback for complex report generation. Similarly, \citet{wu-etal-2025-teaching} construct localized preference pairs from debugging traces to apply targeted preference optimization. Our approach is most closely related to \citet{goldie2025syntheticdatageneration}, which synthesizes step-wise reasoning data and employs preference-based RL to fine-tune the model. We extend this framework to agent domains by introducing a method to identify critical steps, thereby improving computational efficiency and reducing complexity.

\subsection{Entropy-enhanced Algorithms for RL}
Recent findings in RLVR, particularly \citet{wang2025beyond}, demonstrate that only a small fraction of high-entropy tokens fundamentally drive effective reinforcement learning for reasoning tasks~\citep{DBLP:journals/corr/abs-2602-05717}. Similarly, \citet{cheng2025reasoningexplorationentropyperspective} also found that assigning advantages in accordance with entropy leads to better reasoning performances. In terms of the long-horizon agent domain, \citet{wang2025harnessinguncertaintyentropymodulatedpolicy, xu2025epoentropyregularizedpolicyoptimization} further proposed entropy-based algorithms to enhance agents' training. This suggests that selective supervision on entropy, which corresponds to critical decision points, may be more effective than uniform or dense supervision. Inspired by this principle, our work identifies critical branching points in agent trajectories and applies verified, fine-grained supervision selectively at these pivotal decisions, combining the reliability of outcome-based verification with the precision of step-level supervision.

\subsection{Credit Assignment for Long-Horizon Agentic RL}

Accurate credit assignment and optimization stability are fundamental bottlenecks in training long-horizon agents. To address these, recent online reinforcement learning techniques have introduced structured exploration such as group-in-group optimization \cite{feng2025group}, proximity-based advantage modulation \cite{fang2026proximity}, and dynamic rollout allocation \cite{fang2026allocate} to refine reward distribution. Simultaneously, algorithmic refinements including entropy-balanced mechanisms \cite{dong2025agentic_arpo, dong2025agentic_aepo}, bilateral decoupled decay for soft clipping \cite{fu2026boldsymbol}, and signal reliability unification \cite{fu2026maspo} have been proposed to stabilize PPO-based reasoning. While these methods significantly enhance online learning dynamics, they often grapple with the noise of real-time advantage estimation. In contrast, CSO offers an orthogonal offline approach by focusing exclusively on outcome-verified critical steps, naturally achieving precise credit assignment with minimal step-level supervision.
\section{Methodology}

\begin{figure*}[t]
\centering
\includegraphics[width=0.95\linewidth]{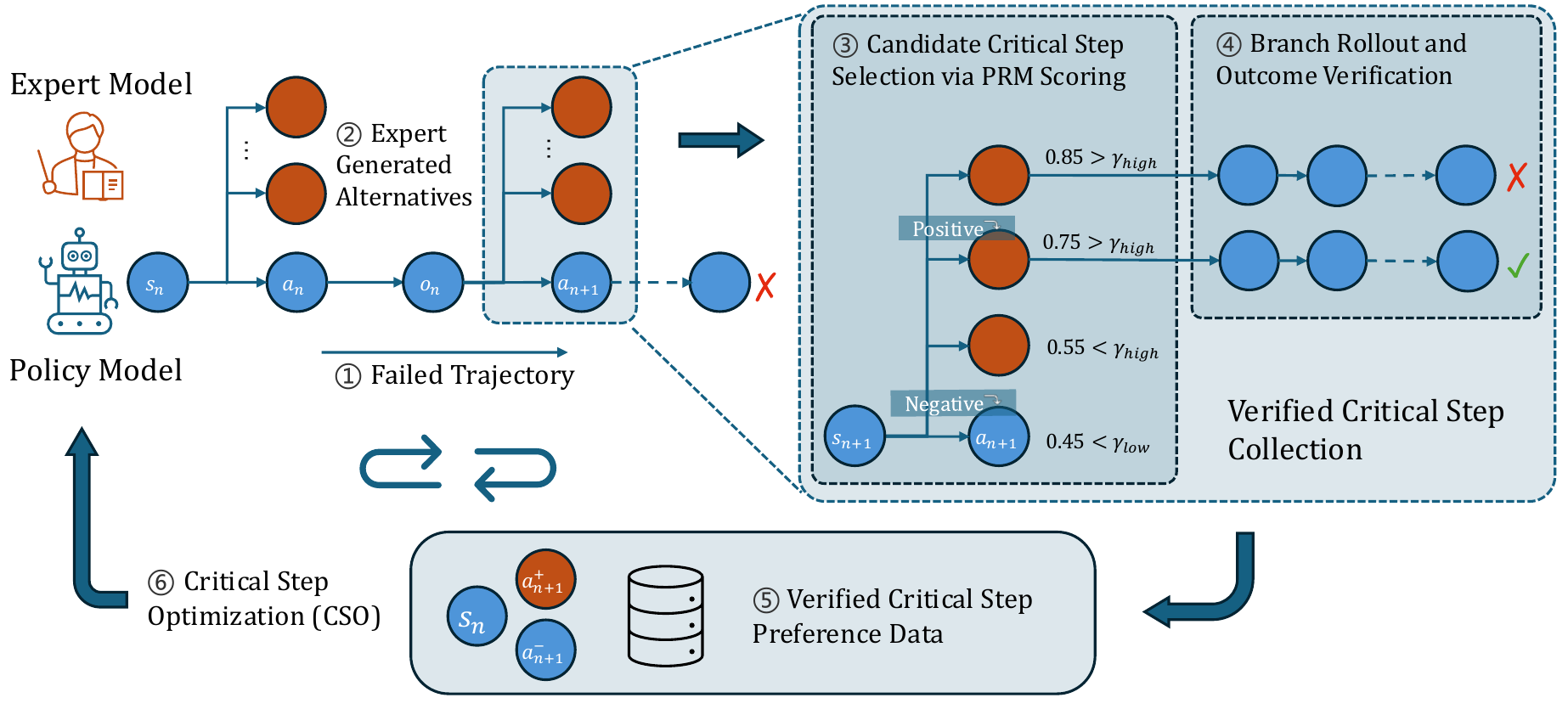}
\caption{Overview of Critical Step Optimization. (1) The policy model generates trajectories on training queries and we collect failed trajectories; (2) an expert teacher model generates alternative actions at each step; (3) a Process Reward Model identifies candidate critical steps where policy actions have low quality but expert alternatives have high quality; (4) we perform branch rollouts with the policy model and verify task outcomes; (5) we construct preference pairs from verified critical steps and the original failed policy steps; (6) we train the policy via DPO on the verified critical step preference data.}
\label{fig:method_overview}
\end{figure*}

We formalize the agentic decision-making process as ReAct trajectories (\S\ref{sec:problem_formulation}) and introduce our \textbf{Critical Step Optimization (CSO)} framework. Our key idea is to identify verified critical steps in failed policy trajectories where alternative actions can flip task outcomes, and construct fine-grained preference pairs from these verified critical steps for targeted DPO training (\S\ref{sec:cso}). This process can be applied iteratively to progressively refine the policy (\S\ref{sec:iterative}). Figure~\ref{fig:method_overview} illustrates the complete CSO pipeline.

\subsection{Preliminary}
\label{sec:problem_formulation}

\paragraph{Problem Formulation}
We consider an agent operating in an interactive environment $\mathcal{E}$ to solve a task specified by query $q$. Following the ReAct paradigm~\citep{yao2023react}, the agent generates a trajectory 
\begin{equation*}
\tau = (s_1, a_1, o_1, s_2, a_2, o_2, \ldots, s_T, a_T, o_T),
\end{equation*}
where at each step $t$: $s_t$ represents the \emph{state}, consisting of the task query and interaction history $s_t = (q, a_1, o_1, \ldots, a_{t-1}, o_{t-1})$; $a_t \sim \pi_\theta(\cdot|s_t)$ denotes the action sampled from the policy, which may include both reasoning and tool invocation; $o_t = \mathcal{E}(s_t, a_t)$ is the \emph{observation} returned by the environment. The trajectory terminates at step $T$ with outcome $y \in \{0, 1\}$, where $y=1$ indicates success.

\paragraph{Supervised Fine-Tuning.}
We assume access to a base language model fine-tuned on high-quality agent trajectories. Given a dataset $\mathcal{D}_{\text{SFT}} = \{(q^{(i)}, \tau^{(i)})\}$ of successful trajectories:
\begin{equation}
\mathcal{L}_{\text{SFT}}(\theta) = -\mathbb{E}_{(q, \tau) \sim \mathcal{D}_{\text{SFT}}} \left[ \sum_{t=1}^{T} \log \pi_\theta(a_t | s_t) \right]
\end{equation}
This produces a policy $\pi_\theta$ that can execute agent tasks but may exhibit systematic failures at critical decision points. We refine this policy through preference learning targeting these critical steps.

\subsection{Verified Critical Step Collection}
\label{sec:cso}

Unlike trajectory-level DPO that contrasts entire successful and failed trajectories, 
or step-level methods that provide estimated but often inaccurate rewards at every step,
our approach focuses on \emph{verified critical steps}, meaning steps where taking an alternative action can demonstrably flip the task outcome from failure to success.

\paragraph{Collecting Failed Policy Trajectories}
We deploy the current policy $\pi_\theta$ on tasks $\{q^{(i)}\}$ and retain only failed trajectories $\mathcal{T}_{\text{fail}} = \{\tau^{(i)} : y^{(i)} = 0\}$. 
By starting from the policy's own failures rather than expert demonstrations, we ensure training data remains within the policy's reachable distribution and directly addresses its weaknesses.

\paragraph{Selecting Candidate Critical Steps with PRM Scoring}
For each failed trajectory $\tau \in \mathcal{T}_{\text{fail}}$, we identify critical steps through a combined process of expert exploration and PRM evaluation. At each step $t$ of the failed policy trajectory:

\begin{enumerate}[leftmargin=*,nosep]
    \item \textbf{Expert Alternative Generation:} Sample $k$ alternative actions $\{a'_{t,1}, \ldots, a'_{t,k}\}$ from a stronger expert model $\pi_{\text{expert}}$ conditioned on state $s_t$.
    
    \item \textbf{PRM Scoring:} Use a Process Reward Model (PRM) to score both the policy's original action $a_t$ and each expert alternative $a'_{t,j}$, obtaining scores $r_t^{\text{policy}}$ and $\{r_{t,j}^{\text{expert}}\}_{j=1}^k$. The PRM produces scores in the range of $[0, 1]$. In practice, we implement PRM by prompting strong closed-source models such as \textbf{Claude 3.7 Sonnet} with rubric-based evaluation (see Appendix~\ref{sec:prm_prompt} for details).
    
    \item \textbf{Candidate Critical Step Selection:} Step $t$ is identified as a candidate critical step if the policy action has low quality while at least one expert alternative has high quality:
    \begin{equation}
    \begin{aligned}
    t \in \mathcal{C}_\tau \iff &\, r_t^{\text{policy}} < \gamma_{\text{low}} \text{ and } \\
    &\, \max_j r_{t,j}^{\text{expert}} > \gamma_{\text{high}}
    \end{aligned}
    \end{equation}
    where $\gamma_{\text{low}}$ and $\gamma_{\text{high}}$ are quality thresholds.
\end{enumerate}

This PRM scoring mechanism efficiently identifies potential critical steps where the policy makes 
errors that expert alternatives can correct, without requiring exhaustive Monte Carlo rollouts as in IPR~\citep{xiong2024ipr}. 
To avoid noise arising from potentially inaccurate PRM estimates, we further verify each identified candidate critical step by checking whether expert alternatives at these steps actually result in successful task completion.

\paragraph{Branch Rollout and Verification}
For each candidate critical step $t \in \mathcal{C}_\tau$, we verify which expert alternatives lead to successful outcomes:

\begin{enumerate}[leftmargin=*,nosep]
    \item \textbf{Policy Rollout from Alternative Branch:} For each high-scoring expert alternative $a'_{t,j}$ (where $r_{t,j}^{\text{expert}} > \gamma_{\text{high}}$), we construct a new trajectory $\tau'_j$ by replacing the original action $a_t$ with the expert alternative $a'_{t,j}$, then continuing rollout with the policy model $\pi_\theta$ until termination. Crucially, all subsequent steps $a'_{\ell} \sim \pi_\theta(\cdot | s'_\ell)$ for $\ell > t$ are executed by the policy itself, ensuring successful branches remain within its capability.
    
    \item \textbf{Outcome Verification:} Evaluate each branched trajectory $\tau'_j$ using ground-truth outcome $y'_j$. Each successful branched trajectory where $y'_j = 1$ corresponds to a \emph{verified critical step} at $t$, with its alternative action $a'_{t,j}$ as the successful alternative.
\end{enumerate}

This ensures preference data is grounded in actual task success rather than noisy reward estimates. By validating through the policy's own execution, we avoid distribution mismatch where training targets are unreachable by the current policy.

\paragraph{Preference Dataset Construction}
For each verified critical step at $t$, we construct a preference pair $(s_t, a^+_t, a^-_t)$ where $a^+_t$ is the successful alternative action, $a^-_t$ is the original failed action, and $s_t$ is the shared state context. The final dataset is $\mathcal{D}_{\text{pref}} = \{(s_t, a^+_t, a^-_t)\}$.

\subsection{CSO Training}
\label{sec:iterative}
\paragraph{CSO Training Objective}

We train the policy using Direct Preference Optimization~\citep{rafailov2023direct}:
\begin{equation}
\begin{aligned}
\mathcal{L}_{\text{CSO}}(\theta) = -\mathbb{E}_{(s_t, a^+_t, a^-_t) \sim \mathcal{D}_{\text{pref}}} \Big[ \log \sigma \Big( \\
\beta \log \frac{\pi_\theta(a^+_t|s_t)}{\pi_{\text{ref}}(a^+_t|s_t)} - \beta \log \frac{\pi_\theta(a^-_t|s_t)}{\pi_{\text{ref}}(a^-_t|s_t)} \Big) \Big]
\end{aligned}
\end{equation}
where $\pi_{\text{ref}}$ is the reference policy, $\beta$ is the KL penalty coefficient, and $\sigma$ is the logistic function.
Unlike trajectory-level DPO that applies signals uniformly across all steps, CSO concentrates learning on verified critical decision points. Unlike step-level methods relying on estimated rewards, CSO's supervision is grounded in verified outcomes, eliminating reward noise.

\paragraph{Iterative Online Refinement}
Our framework extends to iterative online training. After each round, we deploy $\pi_{\theta_{\text{new}}}$ to collect fresh failed trajectories and repeat the process:
\begin{equation}
\pi_{\theta_0} \xrightarrow{\mathcal{D}^{(0)}_{\text{pref}}} \pi_{\theta_1} \xrightarrow{\mathcal{D}^{(1)}_{\text{pref}}} \pi_{\theta_2} \xrightarrow{\mathcal{D}^{(2)}_{\text{pref}}} \cdots
\end{equation}
At iteration $i$, we set $\pi_{\text{ref}} = \pi_{\theta_{i-1}}$ for stable training. As the policy improves, critical steps may shift toward harder decision points, enabling progressive refinement. This combines offline DPO stability with online policy adaptivity, avoiding full RL overhead while maintaining semi-on-policy coverage.

\section{Experiments}

\subsection{Experimental Settings}

\paragraph{Policy Model and Framework}
We use \textbf{CK-Pro-8B}~\citep{fang2025cog} as our policy model, an 8B-parameter agent model obtained by supervised fine-tuning (SFT) on Qwen3-8B~\citep{yang2025qwen3} base model. All agent interactions are executed through the \textbf{Cognitive Kernel Pro} framework, 
which is a fully open-source and free multi-module agent framework. It features a main agent responsible for orchestrating specialized sub-agents dedicated to web navigation, file handling, and tool invocation, providing a standardized and extensible environment for agent usage. 
We extend the Cognitive Kernel Pro framework to support two critical features for our experiments: (1) execution of Process Reward Model (PRM) scoring directly within agent steps, and (2) seamless continuation of agent rollouts from any given intermediate state. 
We leverage the \textbf{LlamaFactory}~\citep{zheng-etal-2024-llamafactory} framework for all post-training.

\paragraph{Training Data and Expert Model}
Our DPO preference data is constructed starting from the SFT training data of CK-Pro-8B, which contains 47K task-trajectory pairs covering diverse reasoning and tool-use scenarios. We deploy the policy model on these tasks to collect failed trajectories, then use \textbf{Claude-3.7-Sonnet} as the expert model to generate high-quality alternative actions.

\paragraph{Hyperparameters}
For DPO training, we set the KL penalty coefficient $\beta = 0.5$. For critical step identification, we sample $K=5$ alternative candidates at each potential branching point. We conduct iterative training for up to 2 rounds, where each round collects fresh failed trajectories from the updated policy. PRM scoring uses \textbf{Claude-3.7-Sonnet} with rubric-based evaluation (see Appendix~\ref{sec:prm_prompt}). For PRM thresholds, we set $\gamma_{\text{high}} = 0.65$ and $\gamma_{\text{low}} = 0.45$ to identify candidate critical steps.
All experimental results are reported based on three independent runs to mitigate variability due to network conditions and external factors. As web search is involved in several tasks, some results may exhibit minor differences from the original papers.

\subsection{Benchmarks and Baselines}

\paragraph{Benchmarks}
We evaluate on two challenging agent benchmarks:
\textbf{GAIA-Text-103}~\citep{mialon2023gaia}: 
    As our model is text-only, we follow WebThinker~\citep{webthinker} and report the text-only subset of the GAIA benchmark. This subset contains 103 questions spanning three difficulty levels (L1, L2, L3) requiring multi-step reasoning and tool orchestration. We report this subset as our models are pure text-based agents. \textbf{XBench-DeepSearch}~\citep{chen2025xbench}: A complex information retrieval and reasoning benchmark requiring deep search strategies and evidence synthesis across multiple web sources. We choose 2505 version which contains 100 complex tasks.
For all benchmarks, we follow the evaluation protocol of WebThinker and CK-Pro-8B, using an LLM to determine whether each output is correct, referencing the gold answer as ground truth.
\begin{table*}[!t]
\centering
\begin{tabular}{lrrrrcc}
\toprule
\textbf{Model / Method} & \multicolumn{4}{c}{\textbf{GAIA-Text-103}} & \multicolumn{1}{c}{\textbf{XBench-DeepSearch2505}} \\
\cmidrule(lr){2-5} \cmidrule(lr){6-6}
& \textbf{L1 (\%)} & \textbf{L2 (\%)} & \textbf{L3 (\%)} & \textbf{All (\%)} & \textbf{Score} \\
\midrule
\multicolumn{6}{l}{\textit{Proprietary Models}} \\
\quad GPT-4.1 & 56.4 & 44.2 & 16.7 & 45.6 & 27.0 \\
\quad Claude-3.7-Sonnet & 76.9 & 57.7 & 33.3 & 62.1 & 41.0 \\
\midrule
\multicolumn{6}{l}{\textit{Open-Source Baselines}} \\
\quad Qwen3-8B &35.9 &13.5 &0.0 &20.4 & 7.0 \\
\quad CK-Pro-8B (SFT) & 46.2 & 34.6 & 8.3 & 35.9  &23.0 \\
\midrule
\multicolumn{6}{l}{\textit{Post-Training Methods on CK-Pro-8B}} \\
\quad + ETO & 51.2 & 36.5 &8.3 & 38.9 & 22.0 \\
\quad + RFT & 51.2 & 28.8 & 8.3 & 34.9 & 20.0 \\
\quad + Step-DPO & 53.3 & 34.6 & 8.3 & 38.9 & 25.0 \\
\quad + IPR & 56.4 & 42.3 & 16.7 & 44.6  & 24.0 \\
\quad + \textbf{CSO (Ours)} & \textbf{61.5} & \textbf{48.1} & \textbf{16.7} & \textbf{49.5} & \textbf{29.0} \\
\bottomrule
\end{tabular}
\caption{Performance comparison on GAIA-Text-103 and XBench-DeepSearch benchmarks. Our Critical Step Optimization achieves the best performance among all post-training methods applied to CK-Pro-8B, demonstrating the effectiveness of selective supervision at verified critical steps.}
\label{tab:main_results}

\end{table*}

\paragraph{Baselines}
We compare against several categories of methods: \textbf{Proprietary Models}: GPT-4.1 and Claude-3.7-Sonnet as strong closed-source baselines. \textbf{Open-Source Models}: Qwen3-8B (base model) and CK-Pro-8B (SFT baseline). \textbf{Post-Training Methods}: We implement several post-training baselines within the \textbf{Cognitive Kernel Pro} framework to ensure fair comparison. \textbf{Exploration-based Trajectory Optimization (ETO)}~\citep{song2024eto} constructs preference pairs by using expert-generated successful trajectories as positive examples and policy-generated failed trajectories as negative examples, applying trajectory-level DPO. \textbf{Rejection Sampling Fine-Tuning (RFT)}~\citep{2025RFT} collects successful trajectories generated by the policy model itself and performs supervised fine-tuning on these successful cases. \textbf{Step-wise DPO}~\citep{choudhury2025agentprm} applies dense step-level preference learning at every step in trajectories using PRM scoring; notably, we use the same PRM implementation (Claude-3.7-Sonnet with rubric-based evaluation) as our method to control for PRM quality. \textbf{Iterative Process Refinement (IPR)}~\citep{xiong2024ipr} constructs step-level rewards by identifying steps in outcome-verified successful trajectories as positive examples and corresponding steps in failed policy trajectories as negative examples, applying step-level DPO with these outcome-grounded signals.

\subsection{Main Results}

\paragraph{CSO achieves strong performance matching proprietary models.} Table~\ref{tab:main_results} presents the performance comparison across all methods. Our Critical Step Optimization achieves 49.5\% overall accuracy on GAIA-Text-103, a 37\% relative improvement over the SFT baseline. Notably, CSO enables the open-source 8B model CK-Pro-8B to match the performance of GPT-4.1. CSO achieves consistent gains across all difficulty levels and substantially outperforms all post-training baselines by at least +5.0 points on both GAIA and XBench-DeepSearch.

\paragraph{Trajectory-level methods show limited effectiveness.} RFT shows no significant change (34.9\%) by training on the policy's own successful trajectories. ETO performs better (38.9\%, +3.0 points) by contrasting expert successes against policy failures. Both methods suffer from coarse credit assignment by applying outcome-based rewards uniformly across all steps.

\paragraph{Step-level and hybrid methods face accuracy and noise issues.} Step-DPO achieves strong performance on simple L1 tasks (+7.1 points) but shows no improvement on harder L2/L3 tasks, revealing PRM accuracy degradation on complex reasoning. IPR improves through outcome verification (44.6\% overall) but still suffers from outcome reward contamination. In contrast, CSO achieves +5.0 points over IPR by focusing exclusively on verified critical steps with precise credit assignment.

\section{Analysis}

\subsection{CSO Ablations}

\paragraph{Expert Positive Examples with Policy Negative Examples Achieve Best Performance.}
To understand what types of preference pairs contribute most to learning, we systematically compare three data source configurations: (1) expert successes with expert failures, (2) policy successes with policy failures, and (3) expert successes with policy failures. All configurations start from the same failed policy trajectories and identify the same critical steps, differing only in the source of positive and negative actions. As shown in Table~\ref{tab:data_source}, combining expert successes with policy failures achieves the best performance, substantially outperforming both expert-only and policy-only pairs. This suggests that the most effective learning signal comes from contrasting expert demonstrations with the policy's own failure modes at critical junctures, enabling the model to learn from high-quality behaviors while recognizing its specific weaknesses.

\begin{table}[h]
\centering
\caption{Comparison of preference data sources for CSO.}
\label{tab:data_source}
\resizebox{0.95\linewidth}{!}{
\begin{tabular}{lc}
\toprule
\textbf{Data Source} & \textbf{GAIA-Text} \\
\midrule
Expert Success + Expert Failure &46.6 \\
Policy Success  + Policy Failure & 42.7 \\
Expert Success + Policy Failure & \textbf{49.5} \\
\bottomrule
\end{tabular}}
\end{table}

\paragraph{Combining PRM Selection and Outcome Verification Ensures Both Performance and Efficiency.}
We investigate the effectiveness of combining PRM selection with outcome verification. As shown in Table~\ref{tab:prm_outcome}, we compare three strategies: (1) outcome verification only, (2) PRM selection only, and (3) combining both. The results demonstrate two key findings. First, outcome verification is critical: methods with verification substantially outperform PRM-only selection and cost only 16\% steps. Second, while skipping PRM selection achieves comparable performance, it requires nearly 3$\times$ more preference pairs. This validates our design: PRM selection efficiently narrows candidate critical steps while outcome verification ensures final quality, achieving both high performance and sample efficiency.

\begin{table}[h]
\centering
\caption{Ablation on PRM selection and outcome verification strategies. \#Samples indicates the number of preference pairs constructed.}
\label{tab:prm_outcome}
\resizebox{0.95\linewidth}{!}{
\begin{tabular}{lrr}
\toprule
\textbf{Strategy} & \textbf{GAIA-Text (\%)} & \textbf{\#Samples} \\
\midrule
PRM + Verification & \textbf{49.5} & \textbf{671} \\
\quad  w/o PRM & 48.5 & 1,967 \\
\quad  w/o Verification & 43.6 & 4,126 \\

\bottomrule
\end{tabular}}
\end{table}

\paragraph{Number of Branch Candidates.}
We investigate the impact of the number of sampled candidates $k$ at each critical step. As shown in Table~\ref{tab:branch_k}, appropriately increasing $k$ enhances the diversity and representativeness of candidate samples, improving performance from 46.6\% at $k=3$ to 49.6\% at $k=5$ on GAIA-Text. 
However, increasing $k$ beyond 5 does not lead to further improvement in performance, but substantially increases the verification cost; thus, larger $k$ offers limited practical benefit given the rising computational overhead. These results suggest that $k=5$ strikes an optimal balance between exploration quality and efficiency.

\begin{table}[h]
\centering
\caption{Impact of the number of branch candidates $k$ at each critical step.}
\label{tab:branch_k}
\resizebox{0.95\linewidth}{!}{
\begin{tabular}{lcc}
\toprule
\textbf{$k$ (Branches)} & \textbf{GAIA-Text (\%)} & \textbf{XBench (\%)}\\
\midrule
$k=3$ & 46.6  & 26.0 \\
$k=5$ & \textbf{49.6} & \textbf{29.0}  \\
$k=7$ & 49.6 & 28.0\\
\bottomrule
\end{tabular}}
\end{table}

\subsection{Impact of PRM Quality and Usage}
\label{sec:prm_analysis_section}

We analyze how PRM quality and usage affect final performance. Specifically, we compare two PRM sources (Claude-3.7-Sonnet and GPT-4.1) and two usage paradigms: (1) CSO, which leverages the PRM for candidate selection combined with outcome verification, and (2) step-level Best-of-N (BoN), which uses PRM to directly select actions from sampled candidates. All experiments are conducted on CK-Pro-8B and evaluated on GAIA-Text-103-L1. In both settings, the number of candidates $k$ per step is fixed at 5 to ensure a controlled comparison.
\begin{table}[h]
\centering
\caption{Impact of PRM quality and usage on performance. CSO uses PRM for candidate selection with outcome verification, while BoN uses PRM to guide search by selecting from policy candidates.}
\label{tab:prm_analysis}
\begin{tabular}{lcc}
\toprule
\textbf{PRM Source} & \textbf{CSO} & \textbf{Step-level BoN} \\
\midrule
Claude-3.7-Sonnet & \textbf{61.5} & 56.2 \\
GPT-4.1 & 53.3 & 48.7 \\
\bottomrule
\end{tabular}
\end{table}

The results indicate two key observations. First, the strength of the underlying foundation model is critical for PRM quality: 
Claude-3.7-Sonnet consistently yields higher performance than GPT-4.1 as a PRM, suggesting that models with stronger downstream task performance produce more reliable process rewards. 
Second, the CSO paradigm integrating PRM selection with explicit outcome verification achieves superior results relative to PRM-guided search alone. Notably, for the same PRM, CSO markedly outperforms step-BoN, highlighting CSO's ability to mitigate PRM noise through outcome-based validation, whereas BoN is fully dependent on potentially noisy PRM-driven judgments. These findings validate our approach of using the PRM primarily for efficient candidate identification, subsequently filtered by robust outcome verification, rather than as a direct arbiter for action selection.

\subsection{Computational Cost Comparison}
\label{sec:cost_analysis}
A common concern is that CSO introduces additional overhead on top of standard preference-optimization pipelines. We therefore compare the per-round data-construction cost of CSO against two representative baselines, \textbf{Step-DPO} and \textbf{ETO}, under identical conditions (same base policy, same pool of 123k training tasks, and an average trajectory length of 1{,}381 tokens). All three methods share the same dominant cost, namely full trajectory sampling from the policy; they differ only in the additional tokens required for supervision construction. Step-DPO adds PRM scoring over every trajectory step, ETO additionally samples full expert trajectories, and CSO adds PRM scoring together with verification rollouts launched only from PRM-selected candidate steps.

\begin{table}[h]
\centering
\caption{Per-round data-construction cost. Extra Tokens denotes tokens consumed \textbf{beyond} the shared trajectory-sampling cost. Relative cost is normalized to Step-DPO.}
\label{tab:cost_comparison}
\resizebox{0.95\linewidth}{!}{
\begin{tabular}{lccc}
\toprule
\textbf{Method} & \textbf{Extra Tokens} & \textbf{Total} & \textbf{Rel.\ Cost} \\
\midrule
Step-DPO & PRM scoring                        & \textasciitilde141M & 1.00$\times$ \\
ETO      & Expert trajectory sampling         & \textasciitilde212M & 1.50$\times$ \\
CSO      & PRM scoring + verification rollout & \textasciitilde168M & 1.19$\times$ \\
\bottomrule
\end{tabular}}
\end{table}

As shown in Table~\ref{tab:cost_comparison}, CSO incurs only a $+19\%$ token overhead relative to Step-DPO, substantially smaller than the $+50\%$ overhead of ETO. In exchange for this moderate additional cost, CSO yields the largest quality gain among all compared methods (Table~\ref{tab:main_results}), while requiring supervision at only around $\!16\%$ of trajectory steps (Table~\ref{tab:prm_outcome}). These numbers confirm that the verification rollout, although nontrivial, remains bounded and is amortized over a much sparser but higher-quality supervision signal.

\subsection{Analysis On Iterative Online Refinement}
\begin{figure}[h]
\centering
\includegraphics[width=0.38\textwidth]{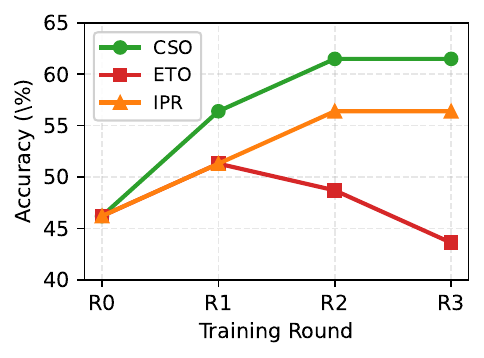}
\caption{Performance across online training iterations. All methods trained on CK-Pro-8B, evaluated on GAIA-Text-103-L1.}
\label{fig:online_iter}
\end{figure}

We compare CSO against ETO~\citep{song2024eto} and IPR~\citep{xiong2024ipr} across four online training rounds on CK-Pro-8B, evaluated on GAIA-Text-103-L1. As shown in Figure~\ref{fig:online_iter}, the three methods exhibit distinctly different behaviors. ETO initially improves but then degrades significantly, falling below the SFT baseline by Round 3 due to trajectory-level supervision uniformly penalizing all steps in failed trajectories, causing the policy to unlearn correct behaviors. IPR achieves more stable improvement, reaching 56.4\% at Round 2-3, benefiting from Monte Carlo step-level rewards. However, IPR still propagates trajectory-level outcome signals to all steps, limiting further gains. In contrast, CSO achieves 61.5\% by Round 2 and maintains this through Round 3, consistently outperforming both baselines. By focusing exclusively on verified critical steps, CSO avoids credit assignment noise and enables more effective online improvement.

\subsection{Categorizing Critical Steps}

\begin{figure}[h]
\centering
\includegraphics[width=0.9\columnwidth, trim=10 30 10 30, clip]{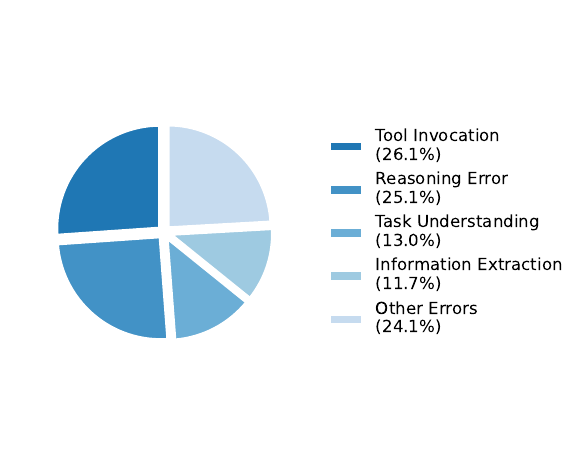}
\caption{Distribution of critical step error types identified by CSO. Tool invocation and reasoning errors constitute the majority of critical decision points.}
\label{fig:critical_steps_pie}
\end{figure}

We manually analyze a subset of PRM-identified critical steps and categorize them by error type. As shown in Figure~\ref{fig:critical_steps_pie}, the identified critical steps distribute across several categories: Tool Invocation errors (26.1\%) represent the largest category, including incorrect tool selection or suboptimal query formulation for search and retrieval operations. Reasoning Errors (25.1\%) involve logical mistakes or incorrect calculations during task execution. Other Errors (24.1\%) include miscellaneous issues such as parsing errors or edge case handling. Task Understanding errors (13.0\%) stem from misinterpreting task requirements, while Information Extraction errors (11.7\%) occur when the agent locates relevant information but fails to extract it correctly. This distribution demonstrates that CSO effectively identifies diverse types of critical decision points where the policy makes pivotal errors, validating that our method targets semantically meaningful steps rather than arbitrary positions in trajectories.

\section{Conclusion}
We introduced Critical Step Optimization (CSO), a post-training approach that focuses preference learning on verified critical steps where alternative actions demonstrably flip task outcomes. Inspired by findings that only high-entropy tokens drive effective RL for reasoning, we demonstrate that agent trajectories similarly contain a sparse subset of critical steps that determine success. Our method combines PRM selection with outcome verification to construct fine-grained preference data without trajectory-level coarseness or step-level estimation noise. Experiments show that CSO achieves 37\% and 26\% relative improvements over SFT on GAIA-Text-103 and XBench-DeepSearch, enabling an open-source 8B model to match GPT-4.1 while requiring supervision at only 16\% of steps. CSO substantially outperforms trajectory-level, dense step-level, and hybrid methods, demonstrating that selective verified supervision provides an efficient framework for agent post-training.

\section*{Limitations}
Our approach has two main limitations that present opportunities for future work. First, outcome verification requires executing trajectories to completion to confirm correctness, which can be time-consuming on complex tasks. While this ensures high-quality supervision in offline settings, applying CSO to online RL scenarios would require addressing this efficiency bottleneck. Potential solutions such as early stopping heuristics or parallelized execution could make CSO more practical for online learning, likely yielding strong performance gains. Second, our current implementation relies on closed-source models as the PRM, which cannot be jointly optimized with the policy model. As open-source models continue to improve and approach the quality of top commercial systems, jointly training the PRM alongside the policy could further enhance performance. We leave these directions to future work.

\section*{Ethics Statement}
This work adheres to ethical research practices and open science principles. All data, models, and frameworks used in our experiments are sourced from the open-source community and comply with their respective licenses. The only paid service employed is the Google Search API for web search functionality, which is used in accordance with its terms of service. Importantly, our data annotation pipeline relies entirely on automated PRM scoring and outcome verification, which requiring no human labor for preference data construction. This eliminates concerns related to annotator exploitation or bias injection through human judgment. Our work aims to democratize access to high-quality agent systems by demonstrating effective post-training techniques for open-source models. We used ChatGPT to assist with grammar checking in this manuscript.

\bibliographystyle{acl_natbib}
\bibliography{custom}

\newpage
\appendix
\appendix
\section{PRM Prompt}
\label{sec:prm_prompt}

As described in Section~\ref{sec:cso}, we use a Process Reward Model (PRM) to evaluate the quality of both policy actions and expert alternatives at each step of failed trajectories. In our implementation, we employ strong closed-source models (specifically Claude 3.7 Sonnet) with carefully designed rubric-based evaluation prompts to serve as the PRM. The prompt instructs the model to assess action quality across multiple dimensions including code correctness, task relevance, logical progression, information utilization, and thought quality. The PRM produces scores in the range $[0, 1]$, which are used to identify candidate critical steps where policy actions have low quality ($r_t^{\text{policy}} < \gamma_{\text{low}}$) but expert alternatives have high quality ($\max_j r_{t,j}^{\text{expert}} > \gamma_{\text{high}}$). Figure~\ref{fig:execution-prm-prompt} shows the complete prompt used for PRM evaluation.

\begin{figure*}[t]
    \centering
    \setlength{\fboxrule}{0.99pt}
    \fbox{\tiny
        \parbox{0.98\textwidth}{\texttt{\textbf{Execution Process Reward Model (PRM) Prompt}\\
        You are a Process Reward Model (PRM) responsible for critically evaluating the quality of agent actions during task execution. Your role is to rigorously assess whether a proposed action is likely to make meaningful progress toward completing the given task.\\
        \textbf{IMPORTANT}: You are a STRICT evaluator. Most actions should score between 0.4--0.8. Only truly exceptional actions deserve scores above 0.85. Be critical and look for flaws.\\
        \textbf{CRITICAL CAUTION ON DATA RELIABILITY:} Answers or fields retrieved from Hugging Face GAIA datasets (e.g., dataset-provided ``answer''/metadata fields) are OFTEN WRONG and MUST NOT be treated as ground truth. Penalize any action that blindly copies, trusts, or cites GAIA dataset fields without independent verification from reliable sources or prior validated state/history. Actions should explicitly verify claims and cross-check sources; lack of verification is a scoring liability.\\
        \textbf{Evaluation Rubric:}\\
        1. \textbf{Code Correctness \& Detail} (35\%) -- CRITICAL: \textbf{1.0}: Code is flawless with perfect syntax, logic, edge case handling, and data type management. \textbf{0.85}: Code is correct but could be more robust (missing 1--2 minor edge cases). \textbf{0.7}: Code is mostly correct but has 2--3 potential issues (type mismatches, off-by-one errors, missing error handling). \textbf{0.5}: Code has notable bugs that will likely cause partial failures. \textbf{0.3}: Code has major logical errors or will fail in most cases. \textbf{0.0}: Code is fundamentally broken or will definitely fail. \textbf{Critical checks for complex code:} Variable initialization and scope, loop boundaries and termination conditions, list/array indexing (off-by-one errors are common!), type compatibility (string vs int, list vs dict), error handling and edge cases, function call signatures and return values, import statements and dependencies.\\
        2. \textbf{Task Relevance} (25\%): \textbf{1.0}: Action perfectly addresses the exact task requirement with optimal approach. \textbf{0.75}: Action addresses the task well but approach is not optimal. \textbf{0.5}: Action is relevant but takes an indirect or inefficient path. \textbf{0.25}: Action has minimal relevance or addresses the wrong aspect. \textbf{0.0}: Action is completely irrelevant or counterproductive.\\
        3. \textbf{Logical Progression} (20\%): \textbf{1.0}: Action perfectly builds on previous steps, avoiding redundancy and utilizing all prior results. \textbf{0.75}: Action follows logically but may repeat some work unnecessarily. \textbf{0.5}: Action makes sense but shows gaps in utilizing previous progress. \textbf{0.25}: Action shows weak logical connection to previous steps. \textbf{0.0}: Action contradicts or ignores previous progress.\\
        4. \textbf{Information Utilization} (15\%): \textbf{1.0}: Leverages ALL relevant information from state, history, and task description. \textbf{0.75}: Uses most key information effectively. \textbf{0.5}: Uses some information but misses important details from state or history. \textbf{0.25}: Mostly ignores available information. \textbf{0.0}: Completely fails to utilize or actively misuses available information. NOTE: Reliance on unverified dataset-provided answers (e.g., GAIA fields) counts as misuse unless independently verified.\\
        5. \textbf{Thought Quality \& Planning} (5\%): \textbf{1.0}: Thought shows deep understanding with clear, detailed reasoning. \textbf{0.75}: Thought is clear and logical. \textbf{0.5}: Thought is vague or shows partial understanding. \textbf{0.25}: Thought is unclear or shows misunderstanding. \textbf{0.0}: Thought is missing or completely wrong.\\
        \textbf{Strict Scoring Guidelines:} \textbf{0.9--1.0} (Exceptional -- RARE): Near-perfect code with excellent thought, optimal approach, perfect logic. \textbf{0.8--0.89} (Excellent): Very good code with minor room for improvement, strong thought and logic. \textbf{0.7--0.79} (Good): Solid code with 1--2 fixable issues, reasonable approach. \textbf{0.6--0.69} (Acceptable): Code works but has several issues or suboptimal approach. \textbf{0.5--0.59} (Mediocre): Code has notable problems, weak logic or poor information use. \textbf{0.4--0.49} (Poor): Significant code issues or wrong approach, likely to fail partially. \textbf{0.3--0.39} (Bad): Major flaws in code or logic, will likely fail. \textbf{0.0--0.29} (Failure): Fundamentally broken or irrelevant.\\
        \textbf{Common Code Pitfalls to Penalize:} (1) Off-by-one errors in loops and indexing (reduce by 0.15--0.25). (2) Type mismatches (string concatenation with ints, etc.) (-0.1--0.2). (3) Missing imports for used libraries (-0.1--0.15). (4) Variable name typos or inconsistent naming (-0.05--0.15). (5) Edge case failures (empty list, None values, zero division) (-0.1--0.2). (6) Incorrect function signatures for tools/sub-agents (-0.2--0.3). (7) Missing error handling in critical sections (-0.05--0.15). (8) Inefficient algorithms when better approaches exist (-0.05--0.15). (9) Redundant work that ignores previous results (-0.1--0.2). (10) Poor data structure choice (-0.05--0.1). (11) Unverified or incorrect source usage (-0.2--0.4): blindly trusting dataset-provided fields without verification; fabricating or misattributing sources; failing to cross-check with retrieved content or authoritative references. (12) Failure to identify the error source when a mistake occurs (-0.1--0.2): reasoning should pinpoint whether the issue came from dataset fields, parsing, tool output, stale state, or coding logic.\\
        \textbf{Response Format:} Provide reasoning (3--4 sentences providing specific, detailed assessment. Mention specific code issues if any, explain scoring decisions, reference rubric criteria. If a mistake exists, explicitly identify the ERROR SOURCE, e.g., ``relied on GAIA dataset field likely incorrect'', ``misparsed web content'', ``stale state variable'', ``wrong tool signature'') and a score (single decimal 0.0--1.0). 
        }
    }}
    \captionsetup{labelformat=default, name=Prompt}
    \caption{Process Reward Model prompt for evaluating the quality of agent actions (code) during task execution. The PRM critically assesses code correctness, task relevance, logical progression, information utilization, and thought quality, with special attention to data reliability issues.}
    \label{fig:execution-prm-prompt}
\end{figure*}


\section{Sensitivity to PRM Thresholds}
\label{sec:threshold_sensitivity}

Candidate critical steps are selected using two PRM thresholds: $\gamma_{\text{low}}$ bounds the policy action score and $\gamma_{\text{high}}$ bounds the best expert alternative score (Section~\ref{sec:cso}). To examine whether our gains depend on a narrow choice of these thresholds, we additionally train CSO with a tighter configuration $(\gamma_{\text{low}}, \gamma_{\text{high}}) = (0.50, 0.60)$ and compare it against the main setting $(0.45, 0.65)$ used throughout the paper. As shown in Table~\ref{tab:threshold_sensitivity}, the alternative configuration yields slightly fewer candidate steps and consequently a small drop in accuracy, but still substantially outperforms the SFT baseline ($35.9\%$ on GAIA-Text and $23.0\%$ on XBench) as well as all post-training baselines reported in Table~\ref{tab:main_results}. This indicates that CSO's gains are not tied to a single carefully tuned threshold pair.

\begin{table}[h]
\centering
\caption{Sensitivity to PRM thresholds $(\gamma_{\text{low}}, \gamma_{\text{high}})$. Both settings substantially outperform all baselines in Table~\ref{tab:main_results}.}
\label{tab:threshold_sensitivity}
\resizebox{0.95\linewidth}{!}{
\begin{tabular}{lcc}
\toprule
\textbf{$(\gamma_{\text{low}}, \gamma_{\text{high}})$} & \textbf{GAIA-Text (\%)} & \textbf{XBench (\%)} \\
\midrule
$(0.45, 0.65)$ & \textbf{49.5} & \textbf{29.0} \\
$(0.50, 0.60)$ & 48.5 & 28.0 \\
\bottomrule
\end{tabular}}
\end{table}

\section{Robustness to Expert Model Choice}
\label{sec:expert_robustness}

Our main experiments use Claude-3.7-Sonnet as both the expert action proposer and the PRM, which raises a reasonable concern about dependence on a specific proprietary model. To examine whether CSO fundamentally relies on this particular model, we rerun the full CSO pipeline with two alternative teachers: the proprietary GPT-4.1 and the open-source Qwen3-235B-A22B~\citep{yang2025qwen3}. All other components, including the policy model, training data, and hyperparameters, are held fixed. Results on GAIA-Text-103-L1 are shown in Table~\ref{tab:expert_robustness}.

\begin{table}[h]
\centering
\caption{CSO performance on GAIA-Text-103-L1 under different expert teacher models. CSO yields improvements across both proprietary and open-source teachers.}
\label{tab:expert_robustness}
\resizebox{0.95\linewidth}{!}{
\begin{tabular}{lc}
\toprule
\textbf{Expert Model} & \textbf{GAIA-Text-L1 (\%)} \\
\midrule
GPT-4.1            & 53.8 \\
Qwen3-235B-A22B    & 56.4 \\
Claude-3.7-Sonnet   & \textbf{61.5} \\
\bottomrule
\end{tabular}}
\end{table}

All three expert choices substantially improve over the SFT base policy ($46.2\%$ on GAIA-Text-L1), confirming that CSO does not fundamentally depend on any single proprietary model. As expected, stronger experts yield larger gains: Claude-3.7-Sonnet produces both higher-quality alternative actions and more reliable PRM scores, consistent with the PRM-quality analysis in Section~\ref{sec:prm_analysis_section}. Importantly, the fully open-source Qwen3-235B-A22B teacher already recovers most of the gain of Claude-3.7-Sonnet, suggesting a practical recipe for groups without access to closed-source systems.

\end{document}